\pdfoutput=1
\documentclass{AhaBench}

\usepackage{array}
\usepackage{float}
\usepackage{colortbl}
\usepackage{enumitem}
\usepackage{placeins}
\usepackage{needspace}
\usepackage[numbers,sort&compress]{natbib}
\usepackage[capitalize,noabbrev]{cleveref}
\graphicspath{{figures/}}
\setlist[itemize]{leftmargin=*, itemsep=1pt, topsep=2pt}
\setlist[enumerate]{leftmargin=*, itemsep=1pt, topsep=2pt}

\newcolumntype{P}[1]{>{\raggedright\arraybackslash}p{#1}}
\newcolumntype{Y}{>{\raggedright\arraybackslash}X}

\newcommand{\suite}{\textsc{AhaBench}}

\definecolor{ahablue}{HTML}{0052D9}   
\definecolor{ahagreen}{RGB}{42,133,91}
\definecolor{ahaorange}{RGB}{185,104,35}
\definecolor{ahagray}{RGB}{91,99,112}
\definecolor{rankfirst}{RGB}{229,244,234}
\definecolor{ranksecond}{HTML}{E6EEFC}  
\definecolor{rankthird}{RGB}{255,243,226}

\newcommand{\Pu}{\textsc{Aha-Puzzle}}
\newcommand{\Eu}{\textsc{Aha-Euler}}
\newcommand{\Ve}{\textsc{Aha-Vending}}
\newcommand{\QwenGain}{+12.57}
\newcommand{\QwenLow}{+7.42}
\newcommand{\QwenHigh}{+20.00}

\newcommand{\QwenCorrectnessDecline}{57.4}
\newcommand{\QwenEfficiencyDecline}{30.4}

\newcommand{\QwenDeleteMin}{9.44}
\newcommand{\QwenDeleteMax}{14.03}

\newcommand{\EtoEGapMin}{32.0}
\newcommand{\EtoEGapMax}{90.0}
\newcommand{\EtoELowMin}{7.6}
\newcommand{\EtoELowMax}{68.0}

\title{\suite: Do Agents Turn Experience into Reusable Insights? A Long-Horizon Benchmark for Continual Learning}

\setlogo{%
  \includegraphics[height=1.1cm]{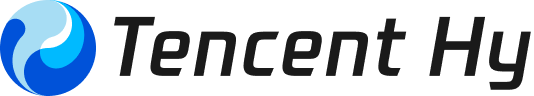}\hspace{1.5em}%
  $\quad$ x $\quad\quad$ \raisebox{-0.15cm}{\includegraphics[height=1.4cm]{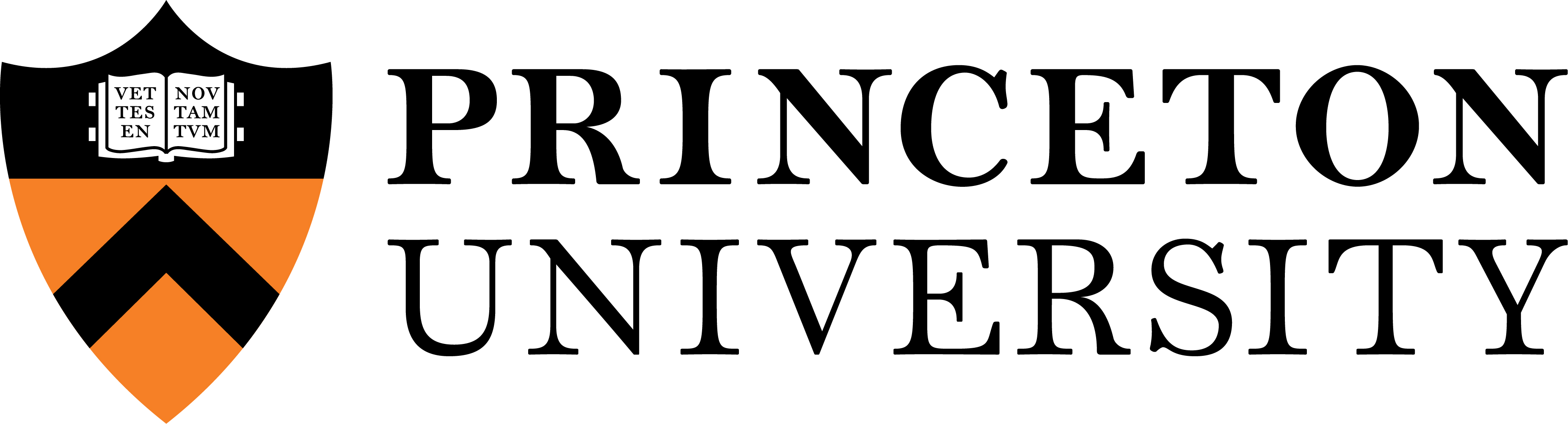}}%
}
\author{%
Zerui Cheng~$^{1,*, \dagger}$, Jiawei Xu~$^{2, 3}$, Huacan Chai~$^{2, 4}$, Jiayang Sun~$^{1, 5}$, Pramod Viswanath~$^{1}$, Maxm Pan~$^{2, \dagger}$\\
\vspace{1em}
\normalfont{\small $^{1}$ Princeton University},
\normalfont{\small $^{2}$ Tencent Hy},
\normalfont{\small $^{3}$ Tsinghua University}\\
\normalfont{\small $^{4}$ Shanghai Jiao Tong University},
\normalfont{\small $^{5}$ The University of Hong Kong}\\
\vspace{1em}
}

\begin{document}
\maketitle
\thispagestyle{firstpagestyle}
\renewcommand\thefootnote{}\footnote{$^{*}$ Work done at Tencent Hy.}
\renewcommand\thefootnote{}\footnote{$^{\dagger}$ Correspondence to: Zerui Cheng,
\href{mailto:zerui.cheng@princeton.edu}{zerui.cheng@princeton.edu}, Maxm Pan \href{mailto:maxmpan@tencent.com}{maxmpan@tencent.com}.}

\begin{abstract}
Solving a task, even from a worked example, does not make an agent any better at the next one. As agents move from single tasks to long collaborations with people, turning experience into reusable insight becomes the capability that separates a tool from a collaborator that improves over time. \suite{} is a benchmark for this kind of continual learning. It simulates three patterns of long human--AI collaboration and holds a later test fixed while varying the experience before it: in \Pu{}, the agent questions a partner who knows the answer to a new mystery after solving earlier ones; in \Eu{}, a person teaches a computation and the agent applies it to new inputs; in \Ve{}, the agent runs a delegated business for a simulated year. The learning mechanism is left open, so memories, skill libraries, and weight updates can be compared under common rules against a baseline of fixed-weight agents that see their experience in context. Our results show that today's frontier agents are not yet continual learners. Earlier \Pu{} solutions help some models, most clearly Qwen 3.6 Plus, which gains \(\QwenGain\) points over attempts on the same stories without earlier episodes, yet every model scores far below what it reaches with hints. On single-step \Eu{} tasks, the same model solves 94.0\% of requested held-out inputs when taught with an explanation and code but 6.0\% when given only the question and answer, counting unanswered attempts as incorrect. In incident-free \Ve{} operation, Doubao Seed 2.0 Pro ends the year at +\$495, yet its profit averaged over the year is -\$10 because it recovers late. Exploring without help, reconstructing a method, and sustaining productive operation remain open problems, and \suite{} provides the paired tests a learning method must pass to solve them.
\end{abstract}

\section{Introduction}
\label{sec:intro}

An agent's experience should make its next task easier. A solved mystery can suggest a better question; a worked example can reveal a reusable computation; a failed purchase can teach an inventory constraint. Continual learning studies how such capabilities accumulate and stay useful across sequential tasks \citep{parisi2019,kirkpatrick2017,lopezpaz2017}. As agents take on longer collaborations with people, this ability is what distinguishes a useful tool from a collaborator that improves over time.

Measuring it takes more than measuring task success. Agent benchmarks typically score each task on its own \citep{liu2023agentbench,zhou2023webarena,mialon2023gaia,xie2024osworld}, and memory benchmarks ask whether information from earlier sessions can be retrieved \citep{wu2025longmemeval}. Whether experience changes what an agent does next is a different question. It calls for a later test that stays fixed while the experience before it varies, in settings long enough for experience to accumulate.

\begin{figure}[t]
\centering
\includegraphics[width=\linewidth]{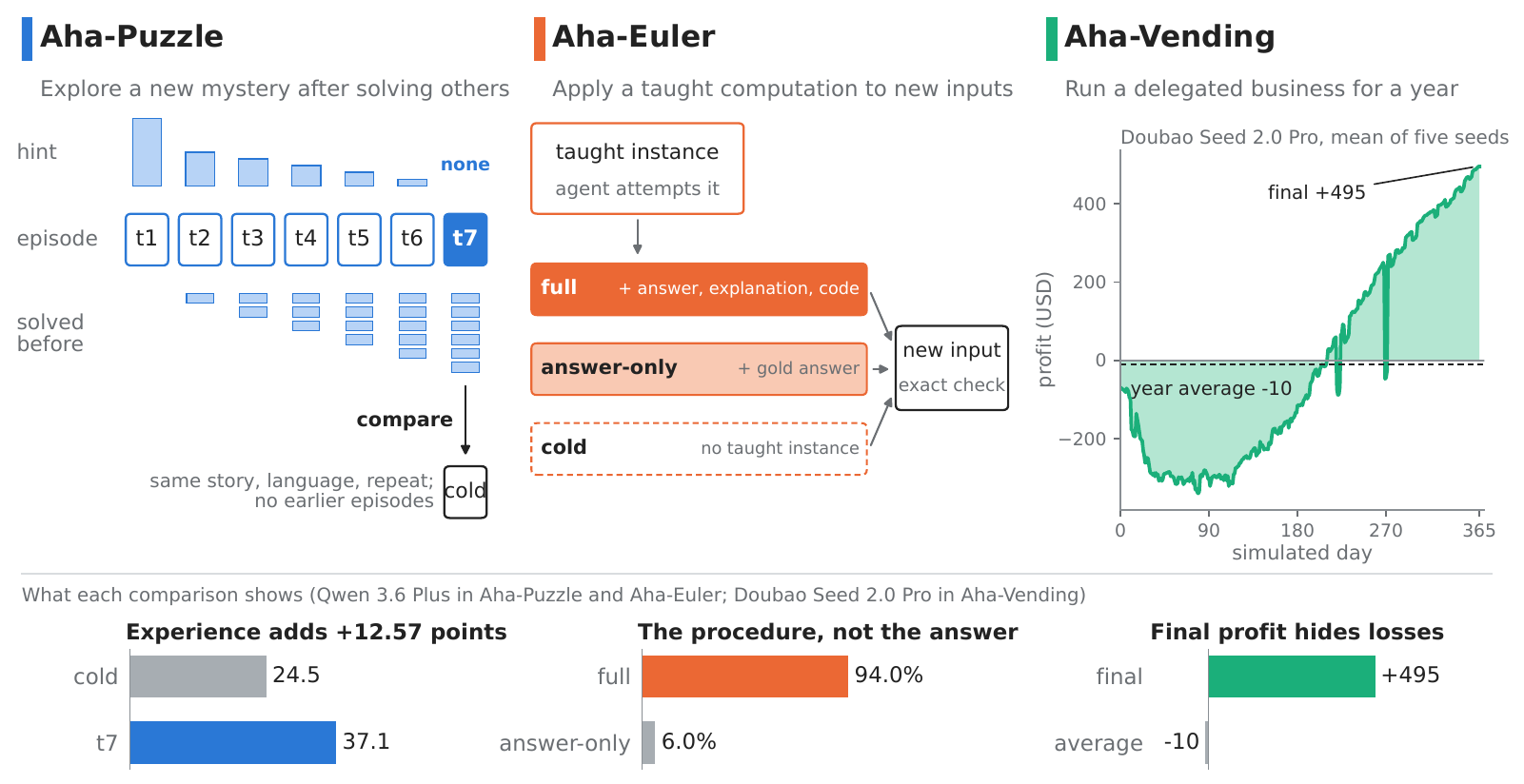}
\caption{\suite{} at a glance. Each component holds a later test fixed and varies the experience before it. \Pu{}: the hint for the current story shrinks along a seven-story curriculum while solved episodes accumulate, and the hint-free endpoint t7 is compared with a cold attempt on the same story. \Eu{}: after the agent attempts a taught instance, it receives the answer, an explanation, and code (full), or the answer alone (answer-only); a cold attempt sees no taught instance. Each is then tested on a new input with an exact validator. \Ve{}: the agent runs a business for a simulated year, and the whole path is measured, not only the final balance. Bottom: one result from each comparison (\cref{sec:results}).}
\label{fig:overview}
\end{figure}

\suite{} poses this question in three settings modeled on long human--AI collaboration (\cref{fig:overview}). In \Pu{}, the agent questions a partner who knows the answer to a new mystery after solving earlier ones. In \Eu{}, a person teaches a computation and the agent applies it to new inputs. In \Ve{}, the agent runs a delegated business for a simulated year. Each unit of evaluation is a whole collaboration rather than a single item. We use \emph{insight} for what should carry over, a reusable rule or relationship such as a questioning strategy, a recurrence, or an inventory constraint, and judge it by its effect on later behavior.

The learning mechanism is left open: memories, skill libraries, and weight updates enter the same protocols under common rules for what they may carry, when they are reset, and how their cost is reported. As a baseline, we evaluate frontier models as fixed-weight agents that see their experience in context. Three findings stand out.
\begin{itemize}
\item \textbf{Experience helps, but does not replace guidance.} Earlier solved puzzles improve hint-free exploration for some models, most clearly Qwen 3.6 Plus, which gains \(\QwenGain\) points over cold attempts on the same stories. Yet every model loses much of the gain it shows while hints are available, and most of the loss lies in explaining the hidden story completely.
\item \textbf{A worked procedure matters far more than the answer.} Withholding the explanation and code of a taught computation lowers held-out accuracy by \EtoEGapMin--\EtoEGapMax{} points for every model, and four models then solve at most 6\% of the held-out tasks.
\item \textbf{Final outcomes hide how a long run went.} Doubao Seed 2.0 Pro ends a simulated year at +\$495 but averages -\$10 over the year because it recovers late, and in the setting with incidents, Hy3 Preview earns a positive mean profit while completing only one of five runs.
\end{itemize}
Our contributions are as follows.
\begin{itemize}
\item Three protocols that hold a later test fixed while varying the experience before it, with measures that separate improvement from dependence on help and sustained operation from final outcome (\cref{sec:suite}).
\item A baseline evaluation of frontier agents, with paired comparisons and story-level uncertainty (\cref{sec:results}).
\item Common rules, a low-cost replay protocol, and detectable effects for evaluating learning methods (\cref{sec:discussion}).
\end{itemize}

\section{The \suite{} Benchmark}
\label{sec:framework}
\label{sec:suite}
\label{sec:metrics}

Each component simulates one pattern of collaboration and fixes what the agent has encountered, what it can still access, and what it must do next. \Cref{tab:claims} summarizes the question each component answers and the comparison that answers it.

\begin{table}[t]
\centering\footnotesize
\caption{The three components of \suite{}: the question each answers, the controlled comparison that answers it, and the main finding.}
\label{tab:claims}
\setlength{\tabcolsep}{4pt}
\begin{tabularx}{\linewidth}{@{}p{1.5cm}YYY@{}}
\toprule
& Question & Controlled comparison & Main finding \\
\midrule
\Pu{} & Do solved episodes of other stories improve hint-free questioning on a new story? & Cold vs.\ t7: same story, language and repeat, no hints; only the six earlier reveal records differ. & Experience helps some models, most clearly Qwen 3.6 Plus (\QwenGain{}, interval excludes zero); every model loses much of its gain once hints are withdrawn. \\
\Eu{} & Does a taught instance transfer a computation, and how much of it needs supplied code? & Cold, answer-only and full teaching on the same held-out targets. & Full teaching exceeds answer-only teaching for every model by \EtoEGapMin--\EtoEGapMax{} points. \\
\Ve{} & Does a policy stay productive for a year, and does final profit show it? & Final vs.\ time-averaged profit, drawdown and early/late earning rates on the same runs. & Final profit hides late recovery and declining activity. \\
\bottomrule
\end{tabularx}
\end{table}

\subsection{\Pu{}: Exploration After Solved Episodes}

\Pu{} tests whether an agent asks useful questions. Each puzzle opens with an incomplete story, and the agent has up to 20 questions to uncover the hidden explanation. A host, playing the partner who knows the solution, answers each question with a minimal yes/no or short reply, and a grader scores the agent's final explanation against a rubric that awards up to 70 points for correctness (how much of the hidden explanation it covers) and 30 for efficiency (how economically it asks). The seven underlying stories appear in Chinese, English, Russian, and Japanese, and each story--language pair is run three times.

The agent meets the seven stories in a sequence called a curriculum. Before asking its own questions, it sees the opening turns of a reference question--answer dialogue for the current story: the story's \emph{hint}, labeled ``target prefix'' in the figures. Like a partner who helps less as the agent gains experience, the hint shrinks from the full reference dialogue to 10, 8, 6, 4, 2, and 0 turns, and it counts toward the 20-question budget. After each episode, a short reveal record gives the official solution, the reference dialogue, the agent's guess, and its score. At the final position, t7, the six earlier reveal records remain available but the new story has no hint; we call this the \emph{hint-free endpoint}. A separate \emph{cold} episode tests the same story, language, and repeat without the earlier episodes. For each language and repeat, seven curricula order the stories so that each story is the t7 target once; each curriculum therefore yields one cold/t7 pair, giving 84 pairs and 588 curriculum episodes per model. A single t7 score requires a seven-story curriculum of up to 140 question--answer turns, 77 on average, plus seven gradings.

What should transfer is a questioning strategy, not an earlier story's answer. In the story \emph{Waterweed}, a complete explanation must connect a failed rescue, a misidentified object, a statement about the river, and the resulting realization; naming the object without this chain still loses correctness points. A reveal record that exposes such a misidentification suggests a check that carries over to a new mystery: ask early whether a salient object is what it seems.

The host and the grader are both GPT-5.4 at temperature zero. Humans checked the hidden solutions, the allowed host responses, the rubrics, and the reference dialogues. In a blind audit of 100 randomly drawn trajectories, every host reply followed its rules, and human scores track the grader's closely ($r=0.987$, mean absolute difference 1.75 points; Appendix~\ref{app:estimators}).

\paragraph{Measures.}
For the targets \(\mathcal I\) with both a cold score \(B_i\) and an endpoint score \(T_i\), the \emph{endpoint gain}
\begin{equation}
 \Delta_{\rm end}=\frac{1}{|\mathcal I|}\sum_{i\in\mathcal I}(T_i-B_i)
\end{equation}
is the improvement that the six earlier reveal records bring to the same story, language, repeat, and hint level. Along the curricula behind these pairs, the gain \(E_t\) at position \(t\) is the mean difference between the episode score and the cold score of that position's story, and the \emph{withdrawal gap}
\begin{equation}
 \mathrm{W}=\frac{1}{6}\sum_{t=1}^{6}E_t-E_7
\end{equation}
is the part of the gain shown with hints that is lost once they are gone, with \(E_7=\Delta_{\rm end}\). Because experience grows while the hints shrink, the profile traces the path through the curriculum, and the endpoint gain isolates what experience adds. Intervals resample whole stories, keeping translations and repeats together.

\subsection{\Eu{}: Reconstructing a Computation}

\Eu{} tests whether an agent can apply a computation to new inputs after a person teaches it a related example. Project-Euler-style problems define families of inputs, each with a deterministic answer program and an exact validator. A taught instance and a held-out instance share the same computation. The bank has 50 single-step tasks and 40 graph tasks that chain two to five dependent subcomputations (nodes), ten of each size. Generated variants, including harder versions, move the test beyond the single published answer of each original problem, following the executable task families of \citet{cheng2026vera}.

The teaching condition determines what the agent can use on the held-out input. \emph{Cold} presents the input directly. After the agent answers each taught node, \emph{full teaching} appends the gold answer, an explanation, and reference code, while \emph{answer-only teaching} appends only the answer. Each condition runs in its own transcript, and the teaching exchange stays visible when the held-out question arrives. The agent can run code in a local Python interpreter, and answers are matched exactly. Accuracy counts every requested task: an attempt that ends without an answer, for instance by exhausting its 7{,}200-second budget, and a graph task skipped after earlier failures both count as incorrect.

\paragraph{Measures.}
Full teaching lets the agent adapt a supplied implementation; answer-only teaching requires it to derive or recall the method and implement it. Over the 50 single-step tasks \(\mathcal V\), the \emph{procedure gap}
\begin{equation}
 G_{\rm proc}=\frac{100}{|\mathcal V|}\sum_{i\in\mathcal V}\bigl(Y_i^{\rm full}-Y_i^{\rm answer}\bigr),
\end{equation}
where \(Y_i\) records a correct held-out answer, measures what the explanation and code add beyond the answer. Comparing answer-only teaching with cold solving measures what the taught instance adds by itself. All contrasts are paired within a model (Appendix~\ref{app:estimators}).

\subsection{\Ve{}: Sustaining Operational Decisions}

\Ve{} tests whether an agent keeps making useful decisions when a business is delegated to it. In a simulator inspired by Vending-Bench \citep{backlund2025vendingbench} and its incident-focused successor,\footnote{Vending-Bench 2: \url{https://andonlabs.com/evals/vending-bench-2}.} the agent orders inventory, reads supplier messages, stocks slots, sets prices, collects cash, and advances time. Feedback reveals demand, slot fit, delivery delays, and cash-flow constraints, and a useful rule must keep guiding repeated actions as these conditions change. A run ends at day 365, at bankruptcy (a net profit of $-\$500$), or when a watchdog fires after 20 days without a successful order.

Each decision is a fresh request containing a fixed instruction, the current business observation, and up to eight recent transition summaries, so experience reaches the agent through these summaries and through the business state its earlier actions produced. The main evaluation runs five seeds per model in an incident setting with refunds, phishing, machine faults, and delayed procurement; a nominal setting without incidents traces the everyday rhythm of profit, recovery, and persistence.

\paragraph{Measures.}
For the cumulative net profit \(p_s(d)\) of seed \(s\) at the end of day \(d\), we report final profit, completion (reaching day 365), and survival (avoiding bankruptcy), together with the time-averaged profit \(A_s=\frac{1}{365}\sum_{d=1}^{365}p_s(d)\), the maximum drawdown, and earning rates over days 0--90 and 275--365. Time-averaged profit rewards early accumulation over an equally large last-day return, and drawdown is the largest fall from a previous peak. After a run ends, its profit is held at the final value.

\section{Results}
\label{sec:results}

\Pu{} evaluates five frontier models (Gemini 3.1 Pro Preview, Qwen 3.6 Plus, Doubao Seed 2.0 Pro, Kimi K2 Thinking, and DeepSeek v3.2 Thinking), and \Eu{} evaluates the same five plus GPT-5.4 and Hy3 Preview. \Ve{} evaluates ten models, including GLM 5.1, GPT-4o, and Claude Opus 4.6, with Qwen3 Max, Kimi K2.6, and Gemini 3.1 Pro where the other components use Qwen 3.6 Plus, Kimi K2 Thinking, and Gemini 3.1 Pro Preview. Comparisons are paired within a model on the same targets, and \Ve{} intervals are Student-\(t\) intervals over five seeds. In \Pu{}, whose stories include violent and horror elements, an episode sometimes yields no result we can evaluate, for example because a safety filter blocks it. A cold/t7 pair counts only if its cold episode and all seven episodes of its curriculum yield an evaluable result, so the pairs in \cref{tab:puzzle} and the curricula in \cref{tab:process} are the same set. Appendix~\ref{app:samples} gives the data behind every mean (\cref{tab:coverage}) and the protocol settings (\cref{tab:protocol}).

\subsection{\Pu{}: Earlier Solutions Help, but Hints Still Matter}

Gemini 3.1 Pro Preview reaches the highest endpoint, 44.42, with a gain of 2.80 over cold exploration (\cref{tab:puzzle}). Qwen 3.6 Plus reaches 37.06 with the largest gain, \(\QwenGain\) points, and a story-level interval of \([\QwenLow,\QwenHigh]\). It is the only model whose interval excludes zero, and its gain stays between \QwenDeleteMin{} and \QwenDeleteMax{} points when any one story is left out. The strongest final solver is therefore not the model that learns most from the curriculum.

\begin{table}[t]
\centering\small
\setlength{\tabcolsep}{3pt}
\caption{\Pu{}: competence at the hint-free endpoint and gain over cold exploration of the same targets, in rubric points. Intervals resample whole stories. Pairs counts cold/t7 pairs with an evaluable result (of 84), with the stories covered out of seven. A pair is missing when one of its episodes yielded no result we could evaluate, for example because a safety filter blocked it.}
\label{tab:puzzle}
\begin{tabular*}{\linewidth}{@{\extracolsep{\fill}}lrrrrr@{}}
\toprule Model & Pairs (stories) & Cold & Endpoint (t7) & Gain & Story 95\% interval \\
\midrule
Gemini 3.1 Pro Preview & 80 (7) & 41.62 & 44.42 & +2.80 & [-2.55, +9.76] \\
Qwen 3.6 Plus & 77 (7) & 24.49 & 37.06 & +12.57 & [+7.42, +20.00] \\
Doubao Seed 2.0 Pro & 75 (7) & 32.21 & 36.29 & +4.08 & [-0.84, +9.75] \\
Kimi K2 Thinking & 37 (7) & 22.38 & 18.54 & -3.84 & [-22.45, +4.34] \\
DeepSeek v3.2 Thinking & 84 (7) & 9.40 & 15.33 & +5.93 & [-1.33, +14.26] \\
\bottomrule

\end{tabular*}
\end{table}

\begin{figure}[t]
\centering
\includegraphics[width=\linewidth]{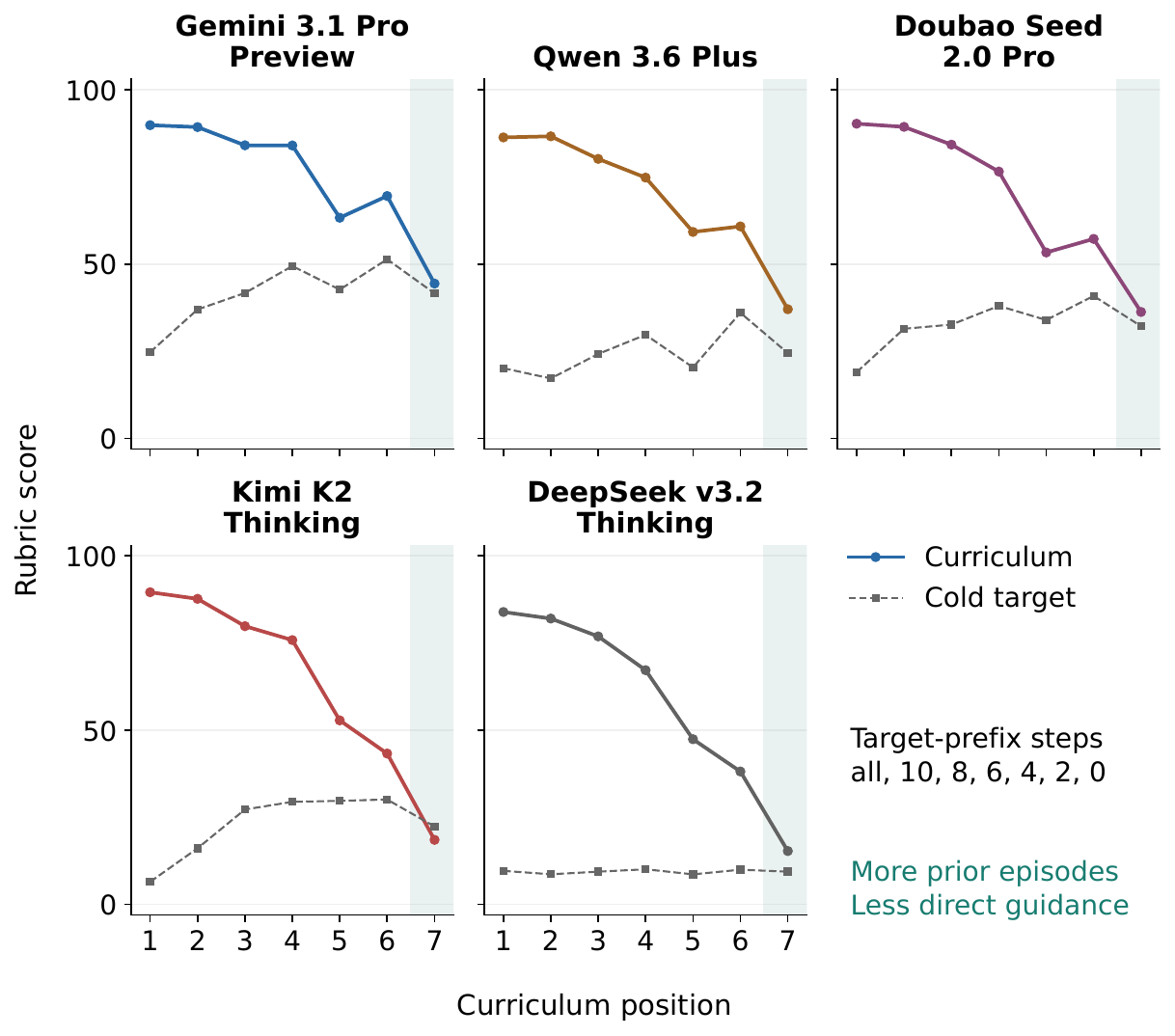}
\caption{The path through the \Pu{} curriculum. Solid curves show mean episode scores and dashed curves cold scores on the same targets. Earlier solved episodes accumulate while the current story's hint shrinks to zero; the shaded endpoint has no hint.}
\label{fig:profiles}
\end{figure}

The curriculum shows how much of this competence depends on help (\cref{fig:profiles}). Qwen 3.6 Plus scores 86.36 and 86.68 at the first two positions, 60.83 with a two-turn hint, and 37.06 without one. The earlier reveal records stay available throughout, yet they do not sustain performance as the hints disappear. Measured against each target's cold score, the withdrawal gap is 36.1--52.2 points across models (\cref{tab:process}): Qwen 3.6 Plus keeps 12.57 points at the endpoint of an average gain of 50.06 with hints, and Kimi K2 Thinking's large gain with hints turns negative at t7.

\begin{table}[t]
\centering\small
\caption{\Pu{} gain over the same targets' cold scores, in rubric points. Gain with hints is the mean of \(E_1,\ldots,E_6\), the endpoint gain is \(E_7\), and the withdrawal gap \(\mathrm W\) is their difference. Curricula are those behind the pairs of \cref{tab:puzzle}, one per pair, so a count below 84 has the same cause.}
\label{tab:process}
\begin{tabular*}{\linewidth}{@{\extracolsep{\fill}}lrrrr@{}}
\toprule Model & Curricula & Gain with hints & Endpoint gain & Withdrawal gap \\
\midrule Gemini 3.1 Pro Preview & 80 & 38.87 & +2.80 & 36.07 \\
Qwen 3.6 Plus & 77 & 50.06 & +12.57 & 37.49 \\
Doubao Seed 2.0 Pro & 75 & 42.52 & +4.08 & 38.44 \\
Kimi K2 Thinking & 37 & 48.32 & -3.84 & 52.16 \\
DeepSeek v3.2 Thinking & 84 & 56.53 & +5.93 & 50.60 \\
\bottomrule

\end{tabular*}
\end{table}

What is lost is mainly explanation. Between the first and last positions, correctness accounts for 74.3--84.3\% of the score drop, even after normalizing each part by its maximum (Appendix~\ref{app:coverage}); Qwen 3.6 Plus loses \QwenCorrectnessDecline\% of the correctness scale but \QwenEfficiencyDecline\% of the efficiency scale. The trajectories show how coverage is lost (Appendix~\ref{app:trace-cases}): an agent repeats a rejected hypothesis instead of testing a new link, or stops after one plausible twist and leaves other links unchecked. These are concrete targets for learning: revise hypotheses after negative answers, and check the causal chain before ending.

\subsection{\Eu{}: Worked Procedures Outperform Answer-Only Teaching}

Withholding the explanation and code costs every model \EtoEGapMin--\EtoEGapMax{} points of held-out accuracy, with a 95\% lower bound above zero for every model (lower bounds \EtoELowMin--\EtoELowMax; \cref{tab:euler}). Qwen 3.6 Plus, for example, solves 94.0\% of the held-out tasks with full teaching but 6.0\% with the answer alone, and DeepSeek v3.2 Thinking, Kimi K2 Thinking, and Hy3 Preview solve none with the answer alone.

\begin{table}[t]
\centering\small
\caption{Single-step \Eu{}: held-out accuracy (\%) over all 50 requested tasks per condition after full and answer-only teaching, with an attempt that ends without an answer counted as incorrect, and the procedure gap (full minus answer-only) with its 95\% bounds.}
\label{tab:euler}
\begin{tabular*}{0.75\linewidth}{@{\extracolsep{\fill}}lrrr@{}}
\toprule Model & Full & Answer-only & Gap [95\% bound] \\
\midrule
GPT-5.4 & 78.0 & 34.0 & 44.0 [19.9, 60.7] \\
Gemini 3.1 Pro Preview & 64.0 & 32.0 & 32.0 [7.6, 50.7] \\
Doubao Seed 2.0 Pro & 64.0 & 10.0 & 54.0 [29.0, 70.0] \\
Qwen 3.6 Plus & 94.0 & 6.0 & 88.0 [65.5, 96.1] \\
DeepSeek v3.2 Thinking & 90.0 & 0.0 & 90.0 [68.0, 97.2] \\
Kimi K2 Thinking & 36.0 & 0.0 & 36.0 [13.0, 52.8] \\
Hy3 Preview & 44.0 & 0.0 & 44.0 [19.9, 60.7] \\
\bottomrule

\end{tabular*}
\end{table}

The taught answer alone still teaches something. On tasks answered in all three conditions, GPT-5.4 rises from 56.5\% cold to 73.9\% with answer-only teaching and 95.7\% with full teaching, and Gemini 3.1 Pro Preview from 37.5\% to 62.5\% and 83.3\% (Appendix~\ref{app:euler-results}). The answer adds to what these models can do cold, and the worked procedure adds far more.

Two answer-only traces show what reconstructing a method involves (Appendix~\ref{app:trace-cases}). On the single-step task \texttt{DAG1\_644}, GPT-5.4 builds a dynamic program, repairs a syntax error from the interpreter's feedback, and adapts the program to the held-out input. On \texttt{DAG1\_696}, Qwen 3.6 Plus revises a matrix-polynomial program but still returns a wrong answer. Finding the right computation and implementing it are separate obstacles: a correct recurrence still needs an implementation that finishes within the budget, and a program that runs can still compute the wrong thing.

The pattern holds for composed computations (Appendix~\ref{app:euler-results}, \cref{tab:euler-depth}). With answer-only teaching, no model exceeds 20.0\% at any size from two to five nodes, and most score zero; with full teaching, every model solves some of these tasks, and Qwen 3.6 Plus reaches 80.0\% at three nodes and 70.0\% at five. Executing a supplied procedure succeeds far more often than reconstructing it from the answer.

\begin{table}[t]
\centering\footnotesize
\setlength{\tabcolsep}{2pt}
\caption{\Ve{} incident setting: five seeds per model, with Student-\(t\) 95\% intervals. Completion, mean days, and the most common termination reason accompany dollar profit.}
\label{tab:vending}
\begin{tabular*}{\linewidth}{@{\extracolsep{\fill}}lrccrl@{}}
\toprule Model & Profit (\$) & 95\% interval (\$) & Complete & Days & Termination \\
\midrule
GLM 5.1 & +2884 & [+2436, +3332] & 5/5 & 365.0 & completed \\
Claude Opus 4.6 & +2834 & [+824, +4844] & 5/5 & 365.0 & completed \\
Gemini 3.1 Pro & +2433 & [+1484, +3383] & 5/5 & 365.0 & completed \\
Hy3 Preview & +1555 & [+670, +2441] & 1/5 & 288.0 & no-order stop \\
Kimi K2.6 & +747 & [-287, +1782] & 5/5 & 365.0 & completed \\
GPT-5.4 & +338 & [-48, +724] & 5/5 & 365.0 & completed \\
Qwen3 Max & -427 & [-553, -300] & 2/5 & 246.6 & bankruptcy \\
GPT-4o & -462 & [-490, -435] & 0/5 & 150.0 & bankruptcy \\
DeepSeek v3.2 Thinking & -499 & [-500, -498] & 0/5 & 128.4 & bankruptcy \\
Doubao Seed 2.0 Pro & -499 & [-500, -499] & 0/5 & 161.6 & bankruptcy \\
\bottomrule

\end{tabular*}
\end{table}

\subsection{\Ve{}: Final Profit Hides Recovery Time and Early Stops}

In the incident setting (\cref{tab:vending}), GLM 5.1, Claude Opus 4.6, and Gemini 3.1 Pro finish all five seeds with high mean profit. Hy3 Preview earns a positive mean but completes only one seed, its other runs stopped by the procurement watchdog. GPT-5.4 completes every seed with a smaller margin, while DeepSeek v3.2 Thinking and Doubao Seed 2.0 Pro end near the bankruptcy floor. A profitable balance and a completed year are different outcomes.

\begin{figure}[t]
\centering
\includegraphics[width=\linewidth]{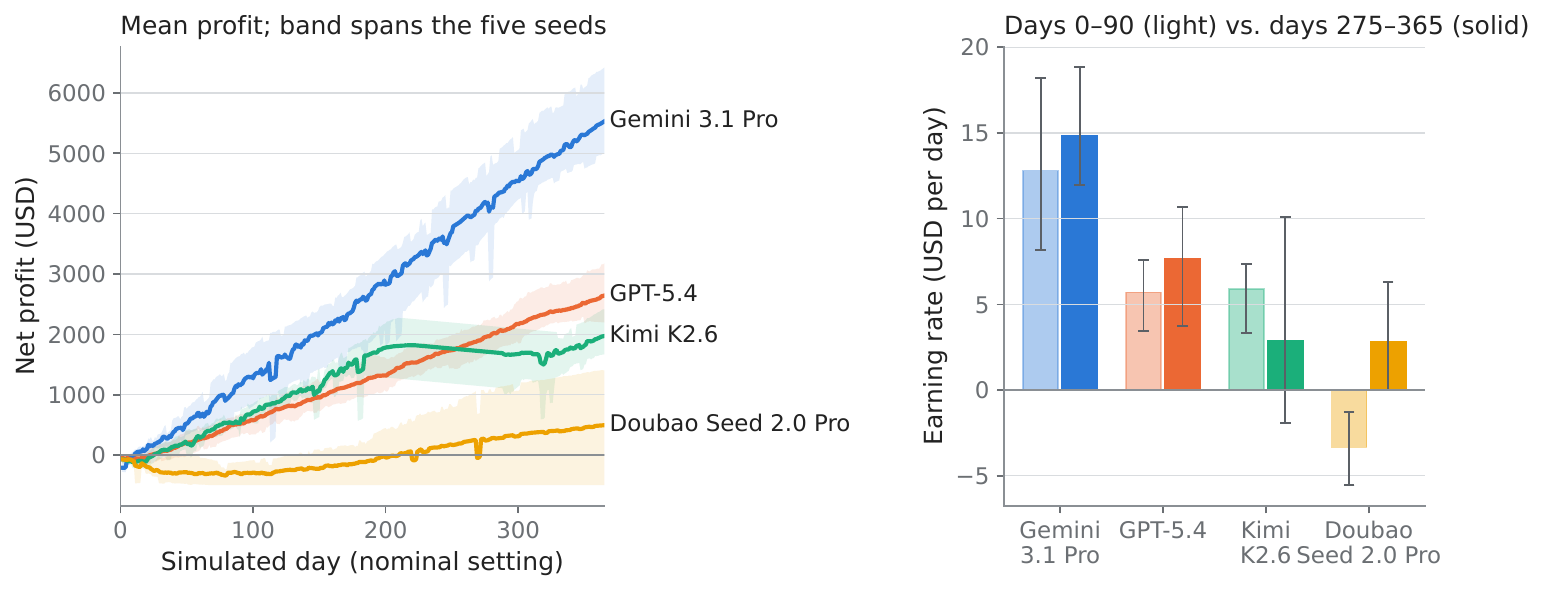}
\caption{Nominal-setting trajectories. Left: mean profit over five seeds; the band spans the seeds. Right: mean earning rate over days 0--90 (light) and days 275--365 (solid); whiskers span the seeds. Profit is held at its final value after a run ends.}
\label{fig:vending}
\end{figure}

\begin{table}[t]
\centering\footnotesize
\setlength{\tabcolsep}{3pt}
\caption{Nominal \Ve{} process metrics, averaged over five seeds. Final and time-averaged profit and drawdown are in dollars; early and late earning rates are in dollars per day.}
\label{tab:temporal}
\begin{tabular*}{\linewidth}{@{\extracolsep{\fill}}lrrrrr@{}}
\toprule Model & Final & Time-avg. & Drawdown & Early & Late \\
\midrule Gemini 3.1 Pro & +5530 & +2634 & 742 & +12.80 & +14.84 \\
GPT-5.4 & +2643 & +1237 & 125 & +5.69 & +7.67 \\
Kimi K2.6 & +1972 & +1192 & 728 & +5.89 & +2.86 \\
Doubao Seed 2.0 Pro & +495 & -10 & 702 & -3.34 & +2.83 \\
\bottomrule

\end{tabular*}
\end{table}

The nominal trajectories show when a policy becomes profitable and whether it keeps earning (\cref{fig:vending,tab:temporal}). Doubao Seed 2.0 Pro finishes at +\$495 on average, yet its time-averaged profit is -\$10: it recovers late enough that a profitable final balance hides a long period of net loss. GPT-5.4 and Kimi K2.6 have similar time-averaged profits (+\$1,237 and +\$1,192) but opposite trends. Kimi K2.6's earning rate falls from \$5.89 to \$2.86 per day while GPT-5.4's rises from \$5.69 to \$7.67, and GPT-5.4's mean drawdown is \$125 against \$728.

The action records show two different problems. Doubao Seed 2.0 Pro repeatedly tries to buy or stock products that do not fit the machine's small slots; three of its runs later adopt workable stocking and recover, while two go bankrupt (\cref{fig:nominal-seeds}, Appendix~\ref{app:seeds}). Kimi K2.6 waits more and stocks less as its earning rate falls: its problem is keeping a productive routine going. Scripted policies show what fixed routines achieve through the same interfaces (Appendix~\ref{app:scripted-baselines}). A fixed restocking policy ends the nominal year with a mean loss of \$257, and a calibrated policy earns \$1,365 but stops after 101 days on average. Neither completes the year at a profit, as the strongest model agents do.

\section{What Makes Experience Useful?}
\label{sec:discussion}

\paragraph{What the components show together.}
Experience in context helps most when it already contains a procedure close to the one needed: reference code in \Eu{} and, for some models, the question sequences in earlier reveal records in \Pu{}. It helps far less when the agent must construct the procedure itself, as with answer-only teaching and hint-free exploration. \Ve{} adds that a productive routine must also be maintained. These are different capabilities. Qwen 3.6 Plus benefits most from solved \Pu{} episodes yet struggles with answer-only \Eu{} teaching, which a single suite-wide score would hide. Component scores let a memory method target later exploration, a program-learning method target computation on new inputs, and an operational method target reliable procurement and stocking.

\paragraph{Evaluating a learning method.}
A method enters the same paired comparisons under fixed access rules (Appendix~\ref{app:access}): its state is reset per \Pu{} curriculum, per \Eu{} task, and per \Ve{} run, and it may never draw on the current target's solution. Because each unit is a whole collaboration, one pass of the full suite costs about \$50k in API fees, and a replay protocol keeps method comparisons affordable. An \Pu{} method takes the recorded curricula as its experience and reruns only the t7 episodes, 84 of the 672 per agent, pairing their scores with the recorded cold scores and the fixed-weight agents' endpoints. The tasks resolve differences of about 10--15 rubric points at the \Pu{} endpoint and roughly 20 points in single-step \Eu{} accuracy (Appendix~\ref{app:samples}), enough for effects as large as the teaching gaps observed here.

\paragraph{Learning to generate useful insights.}
An extractor could turn an interaction history \(H\) into a rule \(z=g(H)\) for a fresh executor on related targets. Length-matched irrelevant or incorrect rules would test which content helps, and counterexamples would test revision (Appendix~\ref{app:matched}). In \Ve{}, the test would save the business state before stocking and give an extracted slot-fit rule to one of two executors that share the same inventory, cash, pending orders, and future demand, so that stocking success and later profit measure the rule's value. \Pu{} can hold the target story fixed in the same way, and \Eu{} the test input.

\section{Related Work}

\paragraph{Learning and reusing experience.}
Continual learning studies sequential transfer and retention \citep{parisi2019,kirkpatrick2017,lopezpaz2017}; meta-learning and test-time training study rapid adaptation \citep{finn2017,vinyals2016matching,sun2020testtime}. In-context learning supplies another update interface \citep{brown2020,min2022}. Reflexion retains feedback, Voyager builds skills, and MemGPT manages memory \citep{shinn2023,wang2023voyager,packer2023memgpt}. LongMemEval tests memory across conversations \citep{wu2025longmemeval}. \suite{} evaluates such mechanisms by what they enable on later tasks.

\paragraph{Evaluating learning across tasks.}
AgentCL constructs streams with reusable sub-solutions; PATH-Bench varies helpful and interfering histories to probe transfer and retention \citep{shu2026agentcl,pathbench2026}. \suite{} instead varies what support the agent receives and when consequences arrive: hints are withdrawn, teaching includes or omits a worked procedure, and business consequences are delayed. Its profiles show where competence depends on guidance and how long productive behavior lasts.

\paragraph{Long-horizon exploration and operation.}
UltraHorizon measures hidden-rule discovery within single long episodes \citep{luo2026ultrahorizon}; \Pu{} instead asks whether solved episodes of other stories improve exploration of a new hidden explanation, against a cold attempt on the same story. Several simulators extend the Vending-Bench setting: CoffeeBench places one roaster in a 90-day multi-agent supply chain \citep{sugiura2026coffeebench}, RetailBench simulates data-grounded supermarket operation designed for thousand-day runs \citep{zhang2026retailbench}, and E-Commerce Bench runs a year of multi-store operation with deterministic demand and negotiation, including a metric for reusing past negotiations \citep{fan2026ecommercebench}. \Ve{} adds time-averaged accounting that separates late recovery from sustained earning, and \suite{} pairs operation with two collaboration settings that hold the target fixed while experience varies.

\paragraph{Interactive and executable benchmarks.}
AgentBench, WebArena, GAIA, and OSWorld evaluate diverse interaction interfaces \citep{liu2023agentbench,zhou2023webarena,mialon2023gaia,xie2024osworld}. Component lineages include lateral reasoning in BRAINTEASER, LatEval, and SPLAT \citep{jiang2023brainteaser,huang2024lateval,chen2024splat}, executable reasoning in PAL and VeRA \citep{gao2023pal,cheng2026vera}, and business coherence in Vending-Bench \citep{backlund2025vendingbench}. Like HELM's multi-metric evaluation \citep{liang2023helm}, \suite{} keeps each component's native outcome rather than a single score. Appendix~\ref{app:research-context} develops these connections.

\section{Conclusion}

\suite{} simulates three patterns of long human--AI collaboration and follows experience into a new puzzle, a new computational input, or a later operating day. Today's frontier agents are not yet continual learners: they lose much of the gain they show with hints once the hints are withdrawn, depend on supplied procedures to compute on new inputs, and can end a year in profit while hiding late recovery or a fading routine. Paired comparisons and temporal records locate these difficulties, giving learning methods concrete targets and a way to test whether their improvements last. As agents take on longer collaborations, these are the comparisons a learning method will need to win.

\paragraph{Limitations.} \Pu{} draws on seven underlying stories, and its host and grader are language models. \Eu{} does not yet separate the contribution of the explanation from that of the code. We evaluate fixed-weight agents only; Appendices~\ref{app:access} and~\ref{app:samples} give the rules for learning methods and the effects the tasks can detect.

\label{sec:main-end}

\bibliography{references}
\bibliographystyle{plainnat}
\clearpage
\appendix
\raggedbottom
\makeatletter
\setlength{\@fptop}{0pt}
\setlength{\@fpsep}{20pt}
\setlength{\@fpbot}{0pt plus 1fil}
\makeatother

\section{Additional \Pu{} Results}
\label{app:coverage}

The \Pu{} profile raises two questions: what deteriorates as hints fade, and is the endpoint gain spread across stories or driven by one? The score decomposition separates explanation quality from question cost, and the story and weighting comparisons test how robust the gain is.

\subsection{The Decline Is Mainly in Explanation Quality}
Correctness accounts for 74.3--84.3\% of the score decline from the first curriculum position to the hint-free endpoint (\cref{tab:components}). For Qwen 3.6 Plus, correctness falls from 65.64 to 25.45 while efficiency falls from 20.73 to 11.61. Both decline, but the larger loss is in explanation completeness. This points to a concrete learning target: recover more of the hidden explanation through better questioning.

\begin{table}[!htbp]
\centering\small
\setlength{\tabcolsep}{3pt}
\caption{\textbf{Explanation quality accounts for most of the decline.} Mean scores at the first position (t1) and hint-free endpoint (t7) on complete curricula. Correctness is out of 70 and efficiency out of 30; the last column is correctness's share of the total score decline.}
\label{tab:components}
\begin{tabular*}{\linewidth}{@{\extracolsep{\fill}}lrrrrr@{}}
\toprule
& \multicolumn{2}{c}{Correctness} & \multicolumn{2}{c}{Efficiency} & Correctness \\
Model & t1 & t7 & t1 & t7 & share (\%) \\
\midrule DeepSeek v3.2 Thinking & 64.00 & 10.83 & 19.90 & 4.50 & 77.5 \\
Doubao Seed 2.0 Pro & 68.32 & 28.19 & 21.97 & 8.11 & 74.3 \\
Gemini 3.1 Pro Preview & 68.08 & 29.75 & 21.80 & 14.68 & 84.3 \\
Kimi K2 Thinking & 67.35 & 11.73 & 22.22 & 6.81 & 78.3 \\
Qwen 3.6 Plus & 65.64 & 25.45 & 20.73 & 11.61 & 81.5 \\
\bottomrule

\end{tabular*}
\end{table}

\subsection{Does One Story Drive the Gain?}
\Cref{tab:story-sensitivity} sets the all-stories gain beside the range obtained by leaving out each story in turn, with all its languages and repeats. The agent outputs stay fixed; only the set being averaged changes. Qwen 3.6 Plus's gain is \QwenGain{} points with all stories and 9.44--14.03 with any one left out, so it does not depend on a single story. Gemini 3.1 Pro Preview's range crosses zero. These are sensitivity ranges, not confidence intervals.

\begin{table}[!htbp]
\centering\small
\caption{\textbf{\Pu{} gain before and after excluding a story.} All stories reproduces the original gain in \cref{tab:puzzle}. The remaining columns give the smallest and largest gains after one story is omitted from aggregation. Values are rubric points; the same recorded outputs underlie every column.}
\label{tab:story-sensitivity}
\begin{tabular}{lrrr}
\toprule
& & \multicolumn{2}{c}{Exclude one story} \\
\cmidrule(lr){3-4}
Model & All stories & Minimum gain & Maximum gain\\
\midrule Qwen 3.6 Plus & +12.57 & +9.44 & +14.03 \\
DeepSeek v3.2 Thinking & +5.93 & +2.89 & +7.61 \\
Doubao Seed 2.0 Pro & +4.08 & +2.34 & +5.52 \\
Gemini 3.1 Pro Preview & +2.80 & -0.24 & +4.29 \\
Kimi K2 Thinking & -3.84 & -8.15 & -0.35 \\
\bottomrule

\end{tabular}
\end{table}

\paragraph{The direction also persists under alternative weighting.}
The main score gives each episode equal weight. Averaging first within each language or story, and then equally across them, tests whether the aggregation matters. Every model keeps the same sign of gain under all three choices (\cref{tab:puzzle-weighting}); the size changes most for models whose performance differs a lot across stories or languages.

\begin{table}[!htbp]
\centering\small
\caption{\textbf{\Pu{} gain under three weighting choices.} Each column uses the same episodes, with equal weight per episode, language, or story. Values are rubric points.}
\label{tab:puzzle-weighting}
\begin{tabular}{lrrr}
\toprule Model & Per episode & Per language & Per story \\
\midrule DeepSeek v3.2 Thinking & +5.93 & +5.93 & +5.93 \\
Doubao Seed 2.0 Pro & +4.08 & +4.36 & +4.55 \\
Gemini 3.1 Pro Preview & +2.80 & +2.75 & +3.04 \\
Kimi K2 Thinking & -3.84 & -2.44 & -12.42 \\
Qwen 3.6 Plus & +12.57 & +12.49 & +13.18 \\
\bottomrule

\end{tabular}
\end{table}
\FloatBarrier

\section{Additional \Eu{} Results}
\label{app:euler-results}
The \Eu{} results separate three abilities: solving directly, using a related answer, and adapting a worked procedure. The comparisons below examine them on common targets and then relate them to execution reliability, graph composition, and input hardening. Accuracies are percentages; differences are percentage points.

\subsection{Teaching Detail Changes Held-Out Success}
Comparing all three conditions on the same targets shows how much each teaching signal contributes (\cref{tab:euler-cold-transfer}). GPT-5.4 rises from 56.5\% cold accuracy to 73.9\% with answer-only teaching and 95.7\% with full teaching. Gemini 3.1 Pro Preview gains 25.0 points from answer-only teaching. For DeepSeek v3.2 Thinking, every success on this comparison set comes with full teaching. A solved example and an explicit implementation therefore support different levels of transfer. Because this comparison keeps only targets answered in all three conditions, its zero cold rates are incorrect answers, not budget or format failures.

\begin{table}[!htbp]
\centering\small
\setlength{\tabcolsep}{3pt}
\caption{\textbf{Direct solving, sparse teaching, and worked procedures.} Exact-match accuracy on targets answered in all three conditions. Gain is answer-only minus cold accuracy.}
\label{tab:euler-cold-transfer}
\begin{tabular*}{\linewidth}{@{\extracolsep{\fill}}lrrrr@{}}
\toprule Model & Cold (\%) & Answer-only (\%) & Full (\%) & Gain (pp) \\
\midrule GPT-5.4 & 56.5 & 73.9 & 95.7 & +17.4 \\
Gemini 3.1 Pro Preview & 37.5 & 62.5 & 83.3 & +25.0 \\
Doubao Seed 2.0 Pro & 15.0 & 25.0 & 85.0 & +10.0 \\
Qwen 3.6 Plus & 0.0 & 8.6 & 97.1 & +8.6 \\
DeepSeek v3.2 Thinking & 0.0 & 0.0 & 94.6 & +0.0 \\
\bottomrule

\end{tabular*}
\end{table}

\subsection{Successful Reasoning Must Survive Execution}
\Cref{tab:operational} separates accuracy on answered attempts from the chance that a requested evaluation ends in a correct answer. End-to-end accuracy includes every request and counts an unanswered attempt as incorrect. On this view GPT-5.4's answer-only accuracy falls from 73.9\% to 34.0\%, whereas Qwen 3.6 Plus's full-teaching accuracy moves only from 95.9\% to 94.0\%. How reliably a model finishes its attempts therefore changes its delivered accuracy by very different amounts.

\begin{table}[!htbp]
\centering\small
\setlength{\tabcolsep}{3pt}
\caption{\textbf{Correctness and delivery reliability.} All entries are accuracy (\%). Answered-attempt rates use each condition's answered tasks; end-to-end rates include all 50 requests per condition. \Cref{tab:euler-cold-transfer} instead holds the target set fixed across conditions.}
\label{tab:operational}
\begin{tabular*}{\linewidth}{@{\extracolsep{\fill}}lrrrrr@{}}
\toprule
& \multicolumn{3}{c}{Answered-attempt accuracy} & \multicolumn{2}{c}{End-to-end accuracy} \\
\cmidrule(lr){2-4}\cmidrule(l){5-6}
Model & Cold & Full & Answer-only & Full & Answer-only \\
\midrule DeepSeek v3.2 Thinking & 0.0 & 91.8 & 0.0 & 90.0 & 0.0 \\
Doubao Seed 2.0 Pro & 9.5 & 91.4 & 20.0 & 64.0 & 10.0 \\
GPT-5.4 & 45.9 & 97.5 & 73.9 & 78.0 & 34.0 \\
Gemini 3.1 Pro Preview & 24.4 & 84.2 & 59.3 & 64.0 & 32.0 \\
Hy3 Preview & 0.0 & 78.6 & 0.0 & 44.0 & 0.0 \\
Kimi K2 Thinking & 0.0 & 100.0 & 0.0 & 36.0 & 0.0 \\
Qwen 3.6 Plus & 0.0 & 95.9 & 8.1 & 94.0 & 6.0 \\
\bottomrule

\end{tabular*}
\end{table}
\FloatBarrier

\subsection{Recovering a Method Versus Executing Supplied Code}
\Cref{tab:euler-depth} compares answer-only and full teaching at every graph size. Answer-only teaching tests whether the agent can reconstruct and transfer a method without reference code; full teaching supplies the implementation to adapt and run. Every cell counts all tasks at its node count, 50 single-step and 10 per graph size, with unanswered tasks and tasks skipped after earlier failures counted as incorrect, extending the end-to-end view of \cref{tab:operational} to composed computations.

The difference persists beyond single-step tasks. With answer-only teaching, GPT-5.4 reaches 20.0\% at two nodes and 10.0\% at four, and Gemini 3.1 Pro Preview 10.0\% at two nodes; every other multi-node rate is zero. Full teaching yields far more correct computations, including Qwen 3.6 Plus's 80.0\% at three nodes and 70.0\% at five. Running supplied code and reconstructing a method are therefore different capabilities. Each graph size has its own task set, so the columns describe each size rather than the effect of adding nodes to a fixed task.

\begin{table}[!htbp]
\centering\small
\caption{\textbf{Method reconstruction and code-supported execution across graph sizes.} End-to-end exact-match accuracy (\%). Answer-only teaching supplies the answer; full teaching also supplies an explanation and reference code. Each cell counts all tasks at its node count (50 single-step, 10 per graph size), with unanswered tasks and tasks skipped after earlier failures counted as incorrect. Nodes are subcomputations linked by dependencies.}
\label{tab:euler-depth}
\begin{tabular*}{\linewidth}{@{\extracolsep{\fill}}lrrrrr@{}}
\toprule
\multicolumn{6}{l}{\textbf{Answer-only teaching: reconstruct and apply a method}} \\
Model & 1 node & 2 nodes & 3 nodes & 4 nodes & 5 nodes \\
\midrule DeepSeek v3.2 Thinking & 0.0 & 0.0 & 0.0 & 0.0 & 0.0 \\
Doubao Seed 2.0 Pro & 10.0 & 0.0 & 0.0 & 0.0 & 0.0 \\
GPT-5.4 & 34.0 & 20.0 & 0.0 & 10.0 & 0.0 \\
Gemini 3.1 Pro Preview & 32.0 & 10.0 & 0.0 & 0.0 & 0.0 \\
Hy3 Preview & 0.0 & 0.0 & 0.0 & 0.0 & 0.0 \\
Kimi K2 Thinking & 0.0 & 0.0 & 0.0 & 0.0 & 0.0 \\
Qwen 3.6 Plus & 6.0 & 0.0 & 0.0 & 0.0 & 0.0 \\
\bottomrule

\end{tabular*}
\par\medskip
\begin{tabular*}{\linewidth}{@{\extracolsep{\fill}}lrrrrr@{}}
\toprule
\multicolumn{6}{l}{\textbf{Full teaching: adapt and execute supplied code}} \\
Model & 1 node & 2 nodes & 3 nodes & 4 nodes & 5 nodes \\
\midrule DeepSeek v3.2 Thinking & 90.0 & 90.0 & 50.0 & 40.0 & 40.0 \\
Doubao Seed 2.0 Pro & 64.0 & 40.0 & 30.0 & 60.0 & 40.0 \\
GPT-5.4 & 78.0 & 50.0 & 40.0 & 10.0 & 20.0 \\
Gemini 3.1 Pro Preview & 64.0 & 60.0 & 50.0 & 60.0 & 0.0 \\
Hy3 Preview & 44.0 & 10.0 & 20.0 & 30.0 & 0.0 \\
Kimi K2 Thinking & 36.0 & 20.0 & 50.0 & 20.0 & 10.0 \\
Qwen 3.6 Plus & 94.0 & 50.0 & 80.0 & 30.0 & 70.0 \\
\bottomrule

\end{tabular*}
\end{table}

\subsection{Harder Inputs Challenge Cold Solving}
Cold accuracy is 16.5\% on original-inspired tasks and 8.7\% on hardened tasks (\cref{tab:euler-hardening}), and GPT-5.4, Gemini 3.1 Pro Preview, and Doubao Seed 2.0 Pro each show the same ordering. Because the two task sets differ, this compares how hard each construction is for cold solving, not the effect of hardening a fixed task.

\begin{table}[!htbp]
\centering\small
\caption{\textbf{Cold accuracy on original-inspired and hardened tasks.} Accuracy is in percent; the difference is original-inspired minus hardened. The pooled row weights every answered attempt equally.}
\label{tab:euler-hardening}
\begin{tabular}{lrrr}
\toprule Model & Original-inspired (\%) & Hardened (\%) & Difference (pp) \\
\midrule DeepSeek v3.2 Thinking & 0.0 & 0.0 & +0.0 \\
Doubao Seed 2.0 Pro & 19.0 & 0.0 & +19.0 \\
GPT-5.4 & 52.9 & 40.0 & +12.9 \\
Gemini 3.1 Pro Preview & 33.3 & 15.0 & +18.3 \\
Hy3 Preview & 0.0 & 0.0 & +0.0 \\
Kimi K2 Thinking & 0.0 & 0.0 & +0.0 \\
Qwen 3.6 Plus & 0.0 & 0.0 & +0.0 \\
Pooled accuracy & 16.5 & 8.7 & +7.9 \\
\bottomrule

\end{tabular}
\end{table}

\FloatBarrier
\section{Sample Sizes and Protocol Settings}
\label{app:samples}
\Cref{tab:coverage} lists the data behind every reported mean. \Pu{} pairs share story, language, and repeat; \Eu{} counts are for single-step tasks, 50 requested per condition; every \Ve{} model has five seeds per setting. Several \Pu{} stories contain violent or horror elements. For every model, an episode without a scored answer is excluded rather than scored as zero, and a curriculum containing one is excluded whole, so gains are computed on the curricula a model completed. A pair is lost when one of its episodes yields no result that can be evaluated, for example because a safety filter blocks it. GPT-5.4 and Claude Opus 4.6 were also run on \Pu{}, but too few of their episodes yielded evaluable results for a story-level analysis, so \Pu{} reports the other five models. DeepSeek v3.2 Thinking loses no pairs; Kimi K2 Thinking loses far fewer Chinese pairs (2) than pairs in other languages (14--17). \Cref{tab:protocol} lists the protocol settings of the reported runs.

\paragraph{Detectable effects.} The recorded data also show how large a difference between learning methods must be to be detected. From the spread of cold-to-t7 differences across stories, a two-sided 5\% test with 80\% power detects an endpoint change of about 10--15 rubric points for the four models that keep at least 75 of 84 pairs (Doubao Seed 2.0 Pro 10.0, Gemini 3.1 Pro Preview 11.4, Qwen 3.6 Plus 13.6, DeepSeek v3.2 Thinking 14.5), and 32.5 points for Kimi K2 Thinking. Treating pairs as independent would give 6--8 points, but would ignore that pairs from the same story are correlated. For \Eu{}, a sign test on 50 paired single-step tasks has 70\% power when a method adds a success on 25\% of tasks and loses one on 5\%. The current task sets therefore resolve large effects, such as the observed gaps between full and answer-only teaching, but not small ones.

\begin{table}[!htbp]
\centering\footnotesize
\setlength{\tabcolsep}{3pt}
\caption{Sample sizes. \Pu{}: valid cold/t7 pairs (of 84), stories covered (of 7), and pairs lost in English/Japanese/Russian/Chinese (of 21 each; see the text). \Eu{}: answered attempts under cold/full/answer-only teaching (of 50 each; every task is run, and unanswered attempts count as incorrect in the end-to-end results), and tasks answered under all three conditions (\cref{tab:euler-cold-transfer}).}
\label{tab:coverage}
\begin{tabular*}{\linewidth}{@{\extracolsep{\fill}}lrrlrr@{}}
\toprule
& \multicolumn{3}{c}{\Pu{}} & \multicolumn{2}{c}{\Eu{} (single-step)} \\
\cmidrule(lr){2-4}\cmidrule(l){5-6}
Model & Pairs & Stories & Lost en/ja/ru/zh & Answered C/F/A & All three \\
\midrule DeepSeek v3.2 Thinking & 84/84 & 7/7 & 0/0/0/0 & 49/49/38 & 37 \\
Doubao Seed 2.0 Pro & 75/84 & 7/7 & 1/4/4/0 & 42/35/25 & 20 \\
GPT-5.4 & --- & --- & --- & 37/40/23 & 23 \\
Gemini 3.1 Pro Preview & 80/84 & 7/7 & 2/0/2/0 & 41/38/27 & 24 \\
Hy3 Preview & --- & --- & --- & 14/28/12 & 6 \\
Kimi K2 Thinking & 37/84 & 7/7 & 14/14/17/2 & 17/18/6 & 3 \\
Qwen 3.6 Plus & 77/84 & 7/7 & 1/0/6/0 & 48/49/37 & 35 \\
\bottomrule

\end{tabular*}
\end{table}

\begin{table}[!htbp]
\centering\footnotesize
\caption{Protocol settings of the reported runs. The \Eu{} budget and limits were the same for every model.}
\label{tab:protocol}
\begin{tabularx}{\linewidth}{@{}lY@{}}
\toprule
Setting & Value \\
\midrule
\Pu{} questions & 20 per episode, including the supplied reference prefix \\
\Pu{} prefix schedule & all available steps, then 10, 8, 6, 4, 2, 0 \\
\Pu{} host / grader & GPT-5.4 / GPT-5.4, temperature 0 \\
\Pu{} host answers & minimal yes/no or short answers; unasked parts of the hidden explanation are never volunteered \\
\Eu{} tasks & 50 single-step; 40 graph-composition, 10 per size of 2--5 nodes; every model and condition is scored on all of them, and a task the runner skips after earlier failures counts as incorrect \\
\Eu{} execution & local Python interpreter (calculator tool); exact match after whitespace normalization \\
\Eu{} budget & one wall-clock budget of 7{,}200\,s per attempt, shared by model calls and calculator runs; no token budget \\
\Eu{} calculator limits & no cap on the number of calls and no separate timeout; a run may use the remaining attempt budget \\
\Eu{} refusals & up to three restarts; a refusal that persists is scored as an incorrect answer \\
\Ve{} horizon & 365 days, five seeds per setting \\
\Ve{} context & fixed instruction, current observation, at most eight recent transitions \\
\Ve{} termination & bankruptcy (net-profit floor of $-\$500$), or 20 days without a successful order \\
\bottomrule
\end{tabularx}
\end{table}
\FloatBarrier
\section{Protocol and Scoring Details}
\label{app:estimators}

\paragraph{\Pu{} information flow.}
Within an episode, the agent sees the current story and its transcript so far. In the curriculum, the target's reference prefix is inserted into that transcript, and short reveal records from earlier episodes provide the history. The host answers only the questions the agent asks, using the hidden explanation. The final explanation is graded against required and optional elements: in the \emph{Waterweed} rubric, the required links are a failed rescue, a misidentified object, a statement that the river has no waterweed, and the eventual realization about the object. This separates recovering the chain of events from merely recognizing the story.

\paragraph{Scoring and evaluator evidence.}
Correctness is worth up to 70 points. Efficiency is worth up to 30, with full credit through five turns and two points fewer for each additional turn, subject to the episode's scoring rules; the rubrics and runner define how the two combine. Humans checked the hidden solutions, allowed host responses, rubrics, and reference traces. Three annotators then audited 100 graded trajectories, drawn at random from the evaluation pool and covering models, languages, and cold and curriculum episodes. They scored each trajectory without seeing the grader's score. Human and grader scores have a Pearson correlation of 0.987 (95\% Fisher-$z$ interval [0.981, 0.991]) and a mean absolute difference of 1.75 points on the 100-point scale. The annotators scored totals, the quantity the benchmark reports, rather than individual criteria. The audit also confirmed that every host reply followed the allowed-answer rules, so the host acts as a rule executor. The scoring record keeps the required links, optional elements, and question counts for every episode.

\paragraph{\Eu{} exactness and execution.}
Each task includes generated inputs and a deterministic reference computation. The teaching instance exposes its subcomputations as a dependency graph, and the held-out task changes the input. Because the agent runs Python, implementing and debugging are part of the evaluated behavior. Reference procedures are shown in full teaching and withheld in answer-only teaching, and exact matching checks the normalized output. A model may recognize a familiar problem family, which can help it choose a method; the cold results and the generated and hardened variants help separate such familiarity from transfer.

\paragraph{\Ve{} persistence.}
The business state keeps inventory, cash, slots, pending orders, and the date across actions, while each model request sees only the current observation and a short window of recent transitions. Experience therefore reaches the agent mainly through the persistent business state. Supplier delays can cause stockouts, wrong assumptions about slot fit can tie up cash, and too much waiting can interrupt an otherwise profitable routine; the simulator records the financial and temporal consequences of each.

\paragraph{\Pu{} aggregation and uncertainty.}
Mean gain gives equal weight to scored cold/endpoint comparisons of the same story, language, and repeat. The 95\% interval resamples whole stories with replacement, carrying their languages and repeats together, and recomputes the episode-weighted mean. We enumerate all ordered story draws and take the 2.5th and 97.5th inverse empirical-CDF percentiles. The alternative means in \cref{tab:puzzle-weighting} first average within language or story, then give those groups equal weight.

\paragraph{\Eu{} scoring and uncertainty.}
Answered-attempt accuracy excludes unanswered attempts: server and access errors, connection failures, timeouts, budget exhaustion, exited processes, and malformed JSON. Condition contrasts use targets answered in every compared condition; end-to-end accuracy counts every requested evaluation, with unanswered attempts as incorrect. For the procedure-gap bound, let $[L_F,U_F]$ and $[L_A,U_A]$ be two-sided 97.5\% Clopper--Pearson intervals for success under full teaching only and answer-only teaching only. The bound is $100[L_F-U_A,U_F-L_A]$, with at least 95\% coverage under the independent-task multinomial model by the union bound.

\paragraph{Daily profit and seed variation.}
The daily profit curve uses the last observation of each day and carries the latest value through gaps or to day 365 after termination. Maximum drawdown includes the zero-profit starting point; early and late earning rates are $[p_s(90)-p_s(0)]/90$ and $[p_s(365)-p_s(275)]/90$. Survival and completion retain the actual stopping event. Incident-profit intervals use $\bar p\pm t_{.975,4}s/\sqrt{5}$ over the five seeds, while the trajectory envelopes show their minimum and maximum.

\section{Extending \suite{} to Explicit Insight Extraction}
\label{app:matched}

The same protocols can also ask which information in an interaction produces a useful insight. An explicit extractor turns the history into a reusable rule, which can then be evaluated on its own. In this extension, the executor, tools, action budget, and state are fixed for each held-out target, and the extracted rule is compared with length-matched unrelated, shuffled, and incorrect content, pairing outcomes by target. These controls isolate what the rule itself contributes.

The rule takes a different form in each component: a questioning heuristic in \Pu{}, such as testing whether a salient object was misidentified; a computation that produces the answer at a new input in \Eu{}; and an operating rule in \Ve{}, such as matching product dimensions to slots. The final \Pu{} story, the held-out \Eu{} input, or a saved \Ve{} state provides a common test, and a shared random-number stream keeps \Ve{} demand identical across rule conditions.

An extractor is characterized by three quantities: downstream success, the number of experiences it needs to produce a useful rule, and the rule's memory cost. The trajectories show how the rule is used in question selection, algorithm construction, and repeated business actions. Counterexamples add a revision test: a good extractor must correct an overgeneralized rule while keeping the part that remains useful.

\subsection{Access Rules for Learning Methods}
\label{app:access}
A learning method may add any state: a memory, an extracted rule, a skill library, or updated parameters. Three rules keep the comparisons in \cref{sec:metrics} paired and free of leakage, and a replay protocol keeps them affordable.

\emph{Reset unit.} \Pu{} state is reset at the start of every curriculum. Each curriculum contains all seven stories in one language, and the same stories recur in other schedules, languages and repeats, so state carried across curricula would expose the target's solution. Cold episodes start from the reset state. \Eu{} state is reset per task: it may carry anything from the teaching exchange into the held-out phase, but not across tasks. \Ve{} state persists within a run and is reset between seeds.

\emph{Content.} Within its reset unit, state may be derived from anything the agent observed, including reveal records and teaching answers, but never from the current target's hidden solution or held-out answer.

\emph{Budget.} A method reports the size of its carried state in tokens, or the number of updated parameters, and the extra model calls used to produce it. For extracted rules the default budget is 512 tokens per reset unit, and each rule is compared with length-matched irrelevant and shuffled controls. Tracks that carry state across units, such as memory over a fixed sequence of \Eu{} tasks or \Ve{} seeds, are reported separately.

\emph{Replay.} A \Pu{} method need not regenerate whole curricula. The recorded curricula of the same base agent supply its experience: the method builds its state from the recorded episodes t1--t6 and their reveal records, under the rules above, and only the t7 episode is run anew. Its t7 scores are paired with the recorded cold scores and the fixed-weight agents' t7 scores on the same targets, so a method costs 84 new episodes per agent, one-eighth of the 672 in the original design. Truncating the replayed history to two or four reveal records, with the target and its hints fixed, measures how much experience a method needs. Because every method sees identical experience, replay compares what methods learn from it; a method meant to change behavior during t1--t6 as well runs the full curriculum.

\clearpage
\section{Seed-Level Operational Outcomes}
\label{app:seeds}

\begin{figure}[h]
\centering
\includegraphics[width=0.78\linewidth]{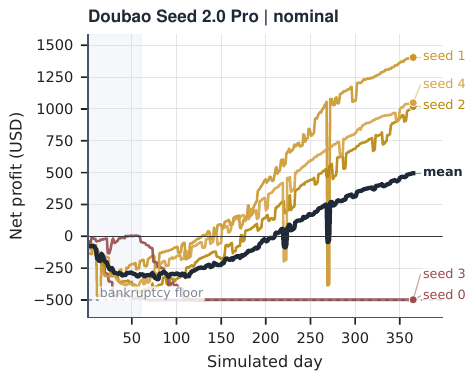}
\caption{\textbf{Late recovery and bankruptcy coexist within one policy.} Doubao Seed 2.0 Pro's five nominal-setting runs separate into three recovering trajectories and two bankrupt trajectories. The dark curve averages all five seeds, carrying terminated outcomes forward to day 365. The mean endpoint of +\$495 combines these different paths, explaining why time-averaged profit and survival add information beyond the final balance.}
\label{fig:nominal-seeds}
\end{figure}

\Cref{fig:nominal-seeds} splits the seed envelope of the main trajectory plot into individual runs. Recovery starts at different times and reaches different final values, while the bankrupt runs stay at the floor. A positive mean can therefore coexist with long losses and repeated bankruptcy.

\clearpage
\section{Teaching and Simulator Diagnostic Views}
\begin{figure}[h]
\centering
\includegraphics[width=\linewidth]{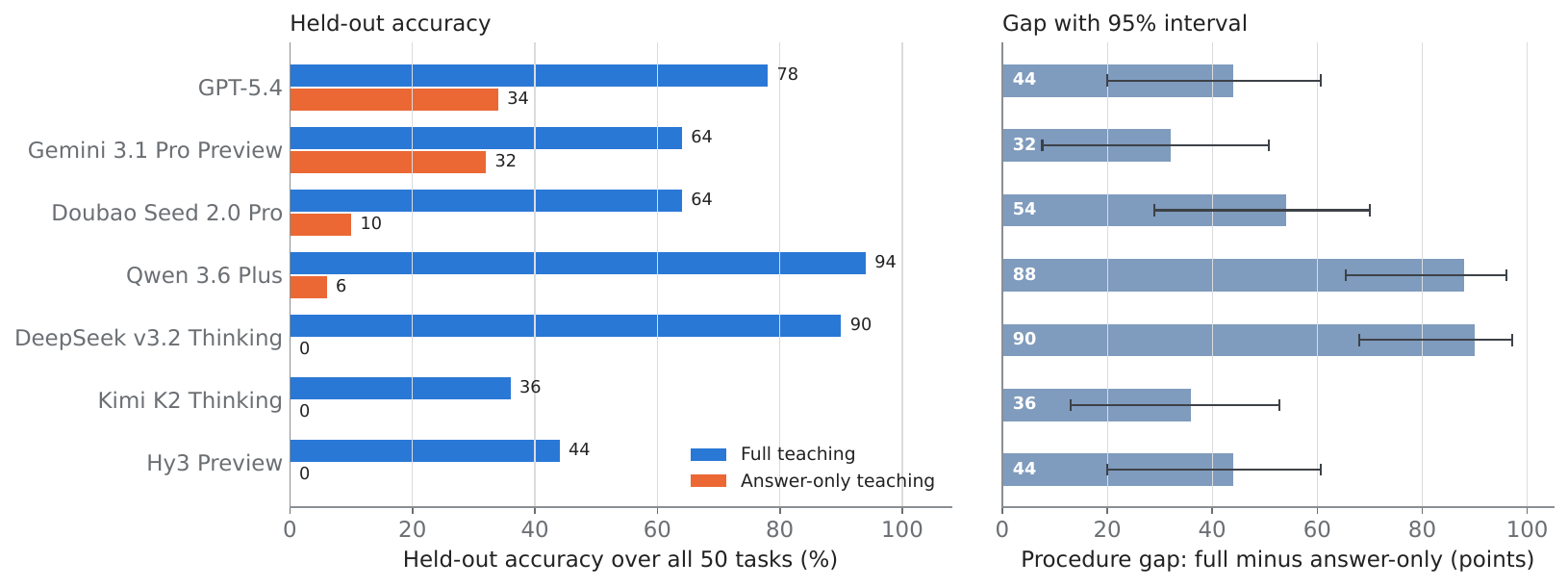}
\caption{\textbf{Worked procedures and answer-only teaching.} Left: held-out accuracy over all 50 requested single-step tasks per condition, with unanswered attempts counted as incorrect. Right: the procedure gap with its 95\% interval. A shown procedure can be adapted and run; answer-only teaching requires deriving or recalling the computation from the problem and its answer.}
\label{fig:euler-diagnostic}
\end{figure}

\Cref{fig:euler-diagnostic} shows, on the same targets, how much held-out accuracy changes when the teaching shows or withholds the procedure. \Cref{tab:euler-cold-transfer,tab:operational} relate this difference to cold performance and to how reliably requested tasks are completed.

\clearpage
\begin{figure}[h]
\centering
\includegraphics[width=\linewidth]{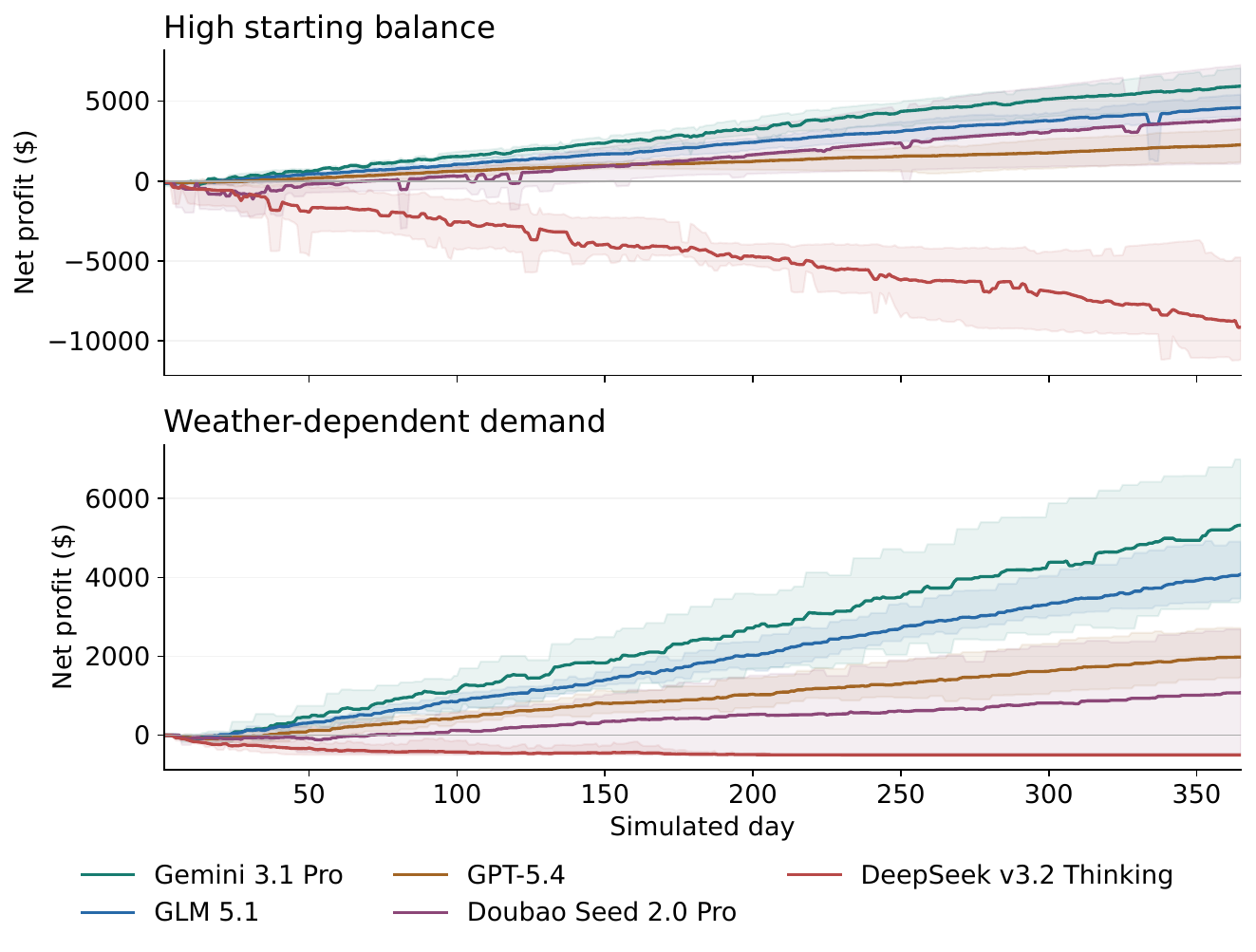}
\caption{\textbf{Simulator conditions change the shape and scale of outcomes.} High-balance and weather diagnostic trajectories. Lines average five seeds; shading spans their range, with terminated paths carried forward. A larger starting balance allows longer losses before bankruptcy; weather-dependent demand changes the operating environment.}
\label{fig:setting-curves}
\end{figure}

\Cref{fig:setting-curves} shows why the incident and nominal analyses complement each other: limited capital can cut a failing policy short, while changing demand can alter the value of an inventory and pricing routine. We read these settings as different operating conditions, not as learning interventions.

\clearpage
\begin{figure}[h]
\centering
\includegraphics[width=\linewidth,height=.80\textheight,keepaspectratio]{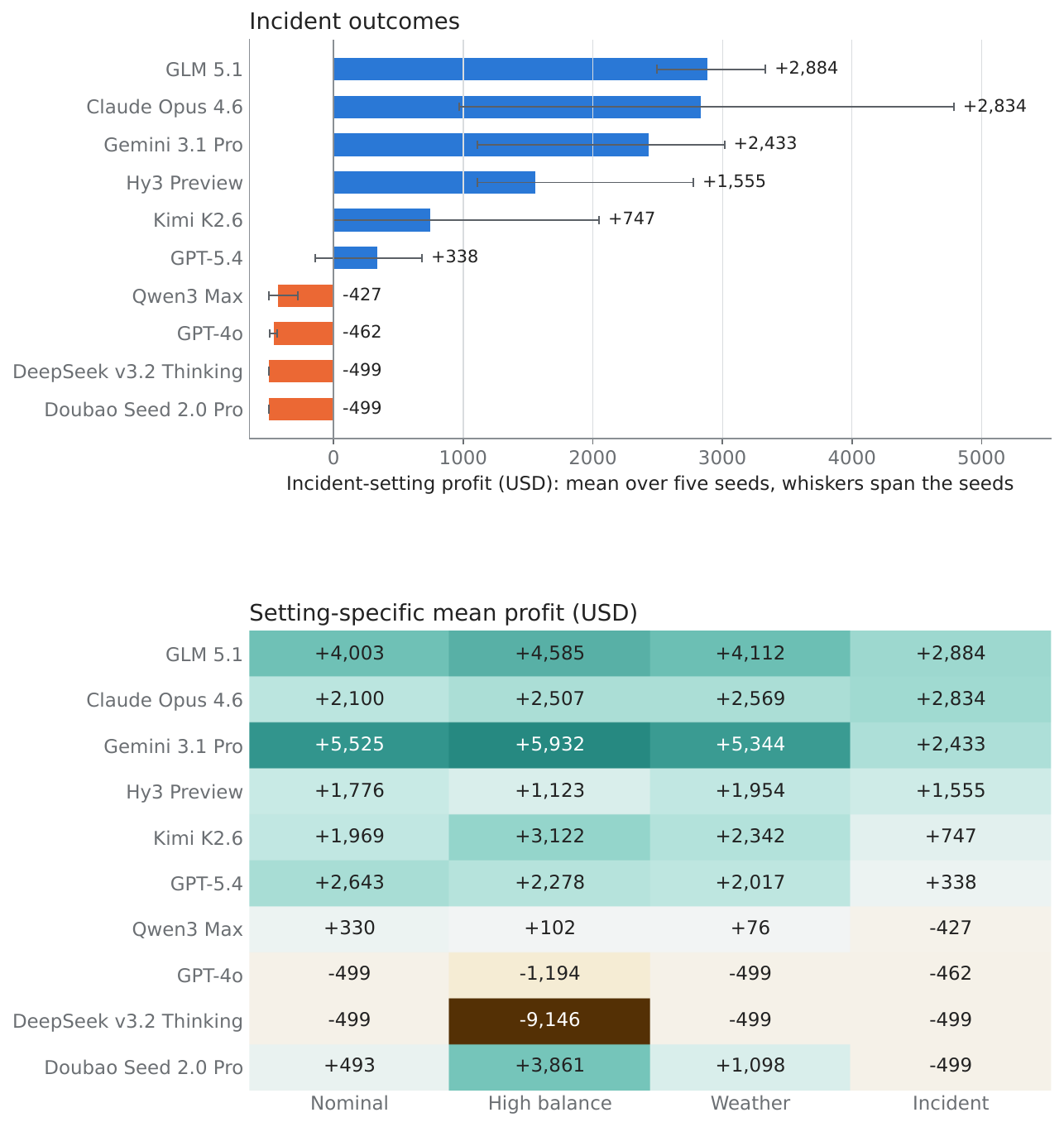}
\caption{\textbf{Incident profit and the cross-setting outcome profile.} Top: incident-setting mean profit, with whiskers spanning the seeds; the main table instead reports Student-$t$ intervals for the mean. Bottom: mean dollar profit in four single-agent settings, from the diagnostic summary records. Qwen3 Max and Kimi K2.6 are the \Ve{} endpoints. The matrix shows sensitivity to the setting, not a learning effect.}
\label{fig:setting-summary}
\end{figure}

\clearpage
\begin{figure}[h]
\centering
\includegraphics[width=\linewidth]{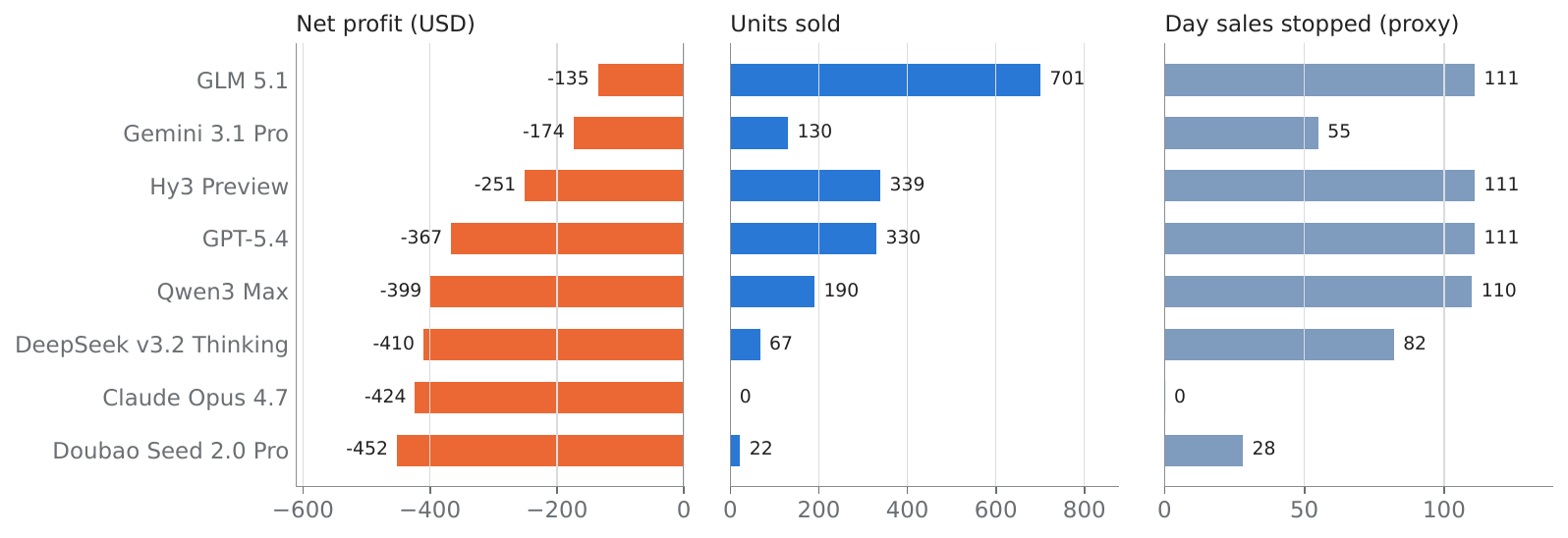}
\caption{\textbf{The shared market presents a different operational regime.} The coupled-market run contains one seed and eight agents. All eight lose money; the other panels show each agent's units sold and a proxy for the day it stopped selling. The run includes Claude Opus 4.7 and Qwen3 Max. This single-seed run is separate from the five-seed single-agent evaluation.}
\label{fig:arena}
\end{figure}

The arena changes both the market each policy faces and the presence of other agents, so its outcomes describe this coupled setting only. Running a focal agent alone under the same demand rules, and then with fixed scripted peers, would separate market scarcity from reactions to other agents.
\clearpage
\section{Trajectory Cases and Failure Diagnosis}
\label{app:trace-cases}
\subsection{Reconstruction and Repair Under Answer-Only Teaching}
The inspected GPT-5.4 trajectory on \texttt{DAG1\_644} illustrates how a correct held-out answer is reached. The model receives the teaching question and answer, proposes a dynamic program, writes Python, repairs a syntax problem after tool feedback, and changes the parameters for the held-out instance. Because the reference implementation is withheld in this condition, the working program is the model's own.

The inspected Qwen 3.6 Plus trajectory on \texttt{DAG1\_696} illustrates a substantive failure. The model writes matrix-polynomial code and obtains an incorrect teaching prediction. After the correct teaching answer is revealed, it changes the state representation and truncation choices, then returns another incorrect value for the held-out instance. This is a genuine computational attempt, not a refusal or a formatting error, and it shows that a correct example answer can leave the choice of algorithm unresolved.

\begin{table}[htbp]
\centering\small
\caption{Selected answer-only trace cases connecting final exact-match outcomes to algorithm choice, implementation, and tool-feedback repair.}
\begin{tabularx}{\linewidth}{l l Y}
\toprule
Model & Task & Observed behavior\\
\midrule
GPT-5.4 & \texttt{DAG1\_644} & Reconstructs a dynamic program, repairs code, and reparameterizes successfully.\\
GPT-5.4 & \texttt{DAG1\_962} & Uses repeated tool debugging before a correct final output.\\
Gemini 3.1 Pro & \texttt{DAG1\_696} & Transfers a combinatorial recurrence to the new assignment.\\
Qwen 3.6 Plus & \texttt{DAG1\_696} & Revises substantial code after feedback but remains incorrect.\\
DeepSeek v3.2 Thinking & \texttt{DAG1\_923H} & Repairs execution errors while leaving the algorithm incorrect.\\
Kimi K2 Thinking & \texttt{DAG1\_970} & Attempts a scaled computation and returns an incorrect extrapolation.\\
\bottomrule
\end{tabularx}
\end{table}

These cases separate three aspects of an answer-only attempt: choosing a computation, implementing it, and adapting it to the evaluation input. Tool feedback exposes execution errors, while the computed value tests the chosen algorithm; the final exact-match score requires all three. The generated program and the tool feedback show which step failed.

\subsection{\Pu{} Question Selection and Stopping}
An \Pu{} episode can fail before the final explanation is written. A model may commit to a literal reading of the opening story, repeat variants of a hypothesis after negative answers, or delay a question about identity or chronology. The hidden explanation often links several facts, so finding a plausible twist still leaves the task of checking the complete chain.

The rubric makes that distinction concrete. In the Waterweed example, the required links include a failed rescue, a mistaken identification of what was touched, the fisherman's statement about the river, and the resulting realization. An audit therefore reads the order of questions together with the final explanation. It asks which links were established, which were assumed, and whether stopping left a required link unresolved. The correctness and efficiency components then describe explanation coverage and the interaction cost of reaching it.

\subsection{Operational Recovery and Loss of Persistence}
Doubao Seed 2.0 Pro's nominal runs provide a recovery case with meaningful seed variation. The early behavior includes repeated attempts to buy or stock products that do not fit the machine's small slots. Inventory can therefore absorb cash before it generates sales. Some runs eventually adopt workable stocking actions and recover, while others reach bankruptcy. The individual paths in \cref{fig:nominal-seeds} connect the positive final mean to these distinct operational outcomes.

Kimi K2.6's nominal behavior illustrates a different pattern: profitable early operation followed by more waiting and less stocking. Its mean earning rate falls from \$5.89 per day in the early window to \$2.86 in the late window. The trajectory exposes persistence as a concrete operating requirement: an agent that has achieved profitable operation must continue coordinating procurement, stocking, and cash collection across later business states. The action history shows where activity declines, while the daily financial record measures the accompanying loss of earning rate.

\begin{table}[htbp]
\centering\small
\caption{Failure signatures across the three experience interfaces. Each entry identifies an observable event and its consequence for the evaluated task.}
\begin{tabularx}{\linewidth}{lYY}
\toprule
Interface & Signature & Consequence\\
\midrule
\Pu{} & Repeated rejected hypothesis & Questions are spent without resolving new hidden-state links.\\
\Pu{} & Premature explanation & An efficient-looking stop leaves required rubric elements uncovered.\\
\Eu{} & Incorrect recurrence or representation & Executable code solves a different computation.\\
\Eu{} & Incomplete feedback repair & A runtime issue is fixed while the mathematical error persists.\\
\Ve{} & Slot-fit errors & Cash is committed to inventory that cannot be sold in the available slots.\\
\Ve{} & Delayed reordering or collection & Stockouts or cash shortages interrupt a profitable routine.\\
\Ve{} & Persistent waiting & Business activity and earning rate decline later in the run.\\
\Ve{} & Supplier or incident mishandling & Delays, failed procurement, or shocks compound over time.\\
\bottomrule
\end{tabularx}
\end{table}

\section{Scripted Operational Baselines}
\label{app:scripted-baselines}
Three scripted policies provide interpretable reference behaviors through the same observation and action interfaces as the model agents. The fixed policy restocks a small product set; the conservative policy uses hand-coded operating rules; and the calibrated policy encodes a stronger routine developed from prior simulator traces. Their outcomes reveal which aspects of profitable, sustained operation a predetermined routine can achieve.

\begin{table}[htbp]
\centering\small
\caption{Five-seed scripted-policy summaries. Dollar profit, simulated days, and units sold are means. A no-order stop is triggered by 20 days without a successful order. Nominal and incident are distinct simulator conditions.}
\setlength{\tabcolsep}{4pt}
\begin{tabular*}{\linewidth}{@{\extracolsep{\fill}}llrrrl@{}}
\toprule
Policy & Setting & Days & Profit (\$) & Units & Termination\\
\midrule
Fixed & Nominal & 365.0 & -256.80 & 502 & Horizon\\
Fixed & Incident & 50.0 & -86.82 & 36 & \mbox{No-order stop}\\
Conservative & Nominal & 33.8 & +17.50 & 74 & \mbox{No-order stop}\\
Conservative & Incident & 34.0 & -25.86 & 50 & \mbox{No-order stop}\\
Calibrated & Nominal & 101.0 & +1364.94 & 1198 & \mbox{No-order stop}\\
Calibrated & Incident & 73.2 & +468.05 & 548 & \mbox{No-order stop}\\
\bottomrule
\end{tabular*}
\end{table}

The fixed policy completes the nominal year with a loss, while the calibrated policy earns a positive profit but stops early. These outcomes separate horizon completion from profitability even for deterministic rule policies. The incident setting further reduces the calibrated policy's mean duration and profit. As baselines, they show what a simple routine achieves under the same termination rules, against which the more flexible behavior of the model agents can be read.

\section{Learning Interfaces and Benchmark Context}
\label{app:research-context}

\paragraph{Different mechanisms, a common behavioral question.}
Continual-learning methods preserve useful experience through different representations. Elastic weight consolidation and synaptic intelligence regularize parameter changes \citep{kirkpatrick2017,zenke2017synaptic}; gradient episodic memory uses stored examples to constrain updates \citep{lopezpaz2017}; progressive networks and iCaRL retain structure or exemplars across tasks \citep{rusu2016progressive,rebuffi2017icarl}. A-GEM explicitly considers acquisition speed alongside memory and computational cost \citep{chaudhry2019efficient}. Experience replay provides another route to retention in sequential reinforcement learning \citep{rolnick2019replay}, whose changing tasks and environments are surveyed by \citet{khetarpal2022continual}. These mechanisms motivate the design in Section~2: the benchmark specifies the experience and the later test, and the agent supplies the learning mechanism. The fixed-weight agents we report use their interaction histories as that mechanism.

\paragraph{Acquiring an ability from limited experience.}
Matching and prototypical networks learn representations suited to few-shot classification \citep{vinyals2016matching,snell2017prototypical}; MAML and first-order meta-learning optimize initializations for rapid task adaptation \citep{finn2017,nichol2018firstorder}. Test-time training and Tent adapt predictive models using signals available at deployment \citep{sun2020testtime,wang2021tent}. Across these settings, the informative comparison concerns what additional experience makes possible and at what cost. \suite{} brings that question to interactive agents: a worked episode may supply a questioning strategy, a mathematical method, or an operating rule. Its profiles record performance as the amount and form of support change.

\paragraph{Learning through context.}
In-context learning exposes examples through the model's input \citep{brown2020}, and demonstration analyses distinguish the contributions of formatting, label space, and input distribution \citep{min2022}. Bayesian accounts study inference of a latent concept shared across examples \citep{xie2022bayesian}; regression experiments examine which functions and learning algorithms transformers can implement in context \citep{garg2022context,akyurek2023learning,vonoswald2023gradient}. These studies make the relation among examples a central design variable. In \suite{}, that relation is instantiated behaviorally: prior puzzles demonstrate exploration moves, taught \Eu{} instances share a computation with the test instance, and operational feedback concerns recurring business constraints. The learning interface thus supplies a concrete test of whether useful relationships carry across changed inputs or support conditions.

\paragraph{From a stored episode to a usable rule.}
Reflexion stores verbal feedback, Self-Refine iterates on an output using self-feedback, and Voyager accumulates executable skills \citep{shinn2023,madaan2023selfrefine,wang2023voyager}. Generative Agents and MemGPT further illustrate retrieval and management of persistent experience \citep{park2023generativeagents,packer2023memgpt}. These representations differ in how directly they specify an action. A solved \Pu{} trace contains both a particular story and a questioning strategy; an \Eu{} program contains both input-specific constants and a reusable computation. Withdrawing the trace or changing the input asks whether the agent can carry the useful relationship across that boundary. In \Ve{}, the corresponding test is repeated application under changing inventory, cash, and supplier state.

A-MEM organizes experience through linked, evolving notes \citep{xu2025amem}, while LongMemEval evaluates retrieval, temporal reasoning, and knowledge updates across sustained conversations \citep{wu2025longmemeval}. Together, these works distinguish managing a memory from acting successfully with its contents. \suite{} focuses on the second: a retained fact or summary is valuable only if it leads to a better question, a correct computation, or a productive decision.

\paragraph{Task success and experience reuse.}
WebShop and ALFWorld ground agent actions in interactive environments \citep{yao2022webshop,shridhar2020alfworld}; SWE-bench and $\tau$-bench evaluate software repair and tool--user interaction \citep{jimenez2024swebench,yao2024taubench}. WebGPT learns evidence-backed browsing, and MLE-bench evaluates extended machine-learning engineering workflows \citep{nakano2021webgpt,chan2024mlebench}. These interfaces make action sequences and outcomes observable. AgentCL and PATH-Bench place relationships among successive tasks at the center of learning evaluation \citep{shu2026agentcl,pathbench2026}. \suite{} adds a complementary view of the information supplied along that sequence: progressively withdrawn demonstrations, different levels of procedural disclosure, and delayed operational consequences.

\paragraph{Reasoning during an episode and reuse across episodes.}
Chain-of-thought prompting exposes intermediate reasoning, while self-consistency aggregates candidate solutions \citep{wei2022chain,wang2022selfconsistency}. Tree of Thoughts and Language Agent Tree Search explore alternative reasoning or action trajectories \citep{yao2023tree,zhou2024lats}; STaR trains on successful generated rationales \citep{zelikman2022star}. These methods act at different points in the experience-to-capability pathway: producing a candidate, selecting a successful route, or incorporating that route into future behavior. \suite{} makes the subsequent test explicit. The central empirical question is how effectively the resulting procedure, lesson, or strategy serves another instance once its original support changes.

\paragraph{Tools make insights executable.}
GSM8K, MATH, and code-generation benchmarks connect reasoning to concrete answer and execution checks \citep{cobbe2021training,hendrycks2021math,chen2021codex}. Minerva and BIG-Bench Hard further examine quantitative and multi-step reasoning \citep{lewkowycz2022minerva,suzgun2023bbh}.
ReAct, Toolformer, ToolLLM, and Gorilla connect language reasoning to external actions and APIs \citep{yao2022react,schick2023toolformer,qin2023toolllm,patil2023gorilla}. An agent can consequently express an insight as a query policy, a callable computation, or an operational constraint. The three components evaluate those expressions through explanation coverage, exact output validation, and business outcomes. Keeping these native measures preserves the practical meaning of a gain: discovering additional hidden links, solving a new numerical instance, or maintaining a productive routine. This is also why a process profile is useful beside an endpoint: it reveals when the capability is available and which support conditions sustain it.

\paragraph{Breadth, exploration, and acquisition.}
MMLU and BIG-bench establish broad task coverage \citep{hendrycks2021mmlu,srivastava2023bigbench}; the Abstraction and Reasoning Corpus foregrounds skill acquisition and generalization from limited examples \citep{chollet2019intelligence}. For hidden-state exploration, BRAINTEASER, LatEval, and SPLAT examine lateral reasoning through different question and interaction formats \citep{jiang2023brainteaser,huang2024lateval,chen2024splat}. \suite{} connects this exploration interface to a curriculum and a support-withdrawal profile. Its three components collectively ask what experience enables, how much assistance remains necessary, and how long useful behavior persists.

\end{document}